\documentclass[letterpaper, 10 pt, journal, twoside]{IEEEtran}

\IEEEoverridecommandlockouts                  

\usepackage{amsmath, amssymb, amsfonts}
\usepackage{graphics} 
\usepackage{graphicx} %
\usepackage{stfloats}
\usepackage{float} %
\usepackage{subfigure} %
\usepackage{threeparttable}
\usepackage{tabularx}
\usepackage{bbding}
\usepackage{pifont}
\usepackage{multirow}
\usepackage{bm}
\usepackage{booktabs}
\usepackage{makecell}  
\usepackage{array}
\usepackage{algorithm}
\usepackage{algpseudocode}
\usepackage[normalem]{ulem}
\usepackage[colorlinks,linkcolor=red]{hyperref}
\usepackage{cite}
\usepackage{balance}

\usepackage{amsmath}

\begin{document}

\markboth{IEEE Robotics and Automation Letters. Preprint Version. Accepted February, 2026}{Wang \MakeLowercase{\textit{et al.}}: WaterSplat-SLAM: Photorealistic Monocular SLAM in Underwater Environmen}

\title{WaterSplat-SLAM: Photorealistic Monocular SLAM in Underwater Environment}

\author{
Kangxu Wang$^{1*}$, 
Shaofeng Zou$^{2*}$, 
Chenxing Jiang$^{3*}$, 
Yixiang Dai$^{1}$, 
Siang Chen$^{1}$, \\
Shaojie Shen$^{3}$, 
Guijin Wang$^{1,\dag}$
 \thanks{Manuscript received: September 30, 2025; Revised December 30, 2025; Accepted February 4, 2026.}
\thanks{This paper was recommended for publication by Editor Civera Javier upon evaluation of the Associate Editor and Reviewers' comments.}
\thanks{$^{1}$Kangxu Wang, Yixiang Dai, Siang Chen and Guijin Wang are with Department of Electronic Engineering, Tsinghua University, Beijing 100084, China
        {\tt\footnotesize \{kx-wang23, daiyx23, csa21\}@mails.tsinghua.edu.cn, wangguijin@tsinghua.edu.cn}}
\thanks{$^{2}$Shaofeng Zou is with Key Laboratory of Marine Robotics, Shenyang 110169, China
University of Chinese Academy of Sciences, Beijing 101408, China
        {\tt\footnotesize zoushaofeng23@mails.ucas.ac.cn}}
\thanks{$^{3}$Chenxing Jiang and Shaojie Shen are with Department of Electronic and Computer Engineering, The Hong Kong University of Science and Technology, Hong Kong, China {\tt\footnotesize cjiangan@connect.ust.hk, eeshaojie@ust.hk}}
\thanks{$^{*}$These authors contributed equally to this work as first authors.}
\thanks{$^{\dagger}$Corresponding Author: {\tt wangguijin@tsinghua.edu.cn.}}
\thanks{Digital Object Identifier (DOI): see top of this page.}
\vspace{-15pt}
}
\maketitle


\begin{abstract}

Underwater monocular SLAM is a challenging problem with applications from autonomous underwater vehicles to marine archaeology. However, existing underwater SLAM methods struggle to produce maps with high-fidelity rendering. In this paper, we propose WaterSplat-SLAM, a novel monocular underwater SLAM system that achieves robust pose estimation and photorealistic dense mapping. 
Specifically, we couple semantic medium filtering into two-view 3D reconstruction prior to enable underwater-adapted camera tracking and depth estimation. Furthermore, we present a semantic-guided rendering and adaptive map management strategy with an online medium-aware Gaussian map, modeling underwater environment in a photorealistic and compact manner. Experiments on multiple underwater datasets demonstrate that WaterSplat-SLAM achieves robust camera tracking and high-fidelity rendering in underwater environments. The code will be realeased at \url{https://github.com/KX-Wang77/WaterSplat-SLAM}.

\end{abstract}

\section{INTRODUCTION}

\IEEEPARstart{O}{line} photorealistic dense mapping is crucial for many underwater robotic applications, such as pipeline inspection, deep-sea exploration, and marine robotics. With real-time, high-fidelity dense mapping, robots can perform advanced tasks and provide precise environmental feedback.

Existing online underwater mapping systems often rely on sonar arrays,~\cite{uw_sensor_survey}, which suffer from low spatial resolution, a lack of photometric information, and high cost. In contrast, vision sensors offer a promising and cost-effective alternative. Prior visual underwater SLAM methods largely relied on handcrafted features ~\cite{evaluation_monoslam,orbslam_underwater,zhao2020detecting}, which often break down under poor illumination, turbidity, and low texture environments. Moreover, feature-based methods typically produce sparse maps that lack the dense, high-fidelity geometry and appearance needed for advanced underwater robotics. To obtain dense maps, recent dense SLAM approaches ~\cite{wang2020acoustic, Hybrid-VINS, ou2025underwater, ou2024structured} fuse structured light, sonar, or laser to build dense point clouds or occupancy maps, but these map representations are not suited for photorealistic rendering.

Recent works~\cite{Matsuki_2024_CVPR, yugay2024gaussianslamphotorealisticdenseslam, huang2024photo, zhang2024hi, lee2024mvsgshighquality3dgaussian} integrate 3D Gaussian Splatting (3DGS) into SLAM frameworks, enabling joint camera poses estimation and online photorealistic mapping. {However, unlike terrestrial environments, underwater scenes suffer from coupled optical degradations such as attenuation and scattering.} These methods do not account for the unique scattering and absorption of water and thus do not model it as a light‑scattering medium, instead using geometric Gaussians that impede underwater planning, grasping, and exploration. 
In addition, the fused representations degrade tracking stability and accuracy in underwater environments.

To overcome these challenges, we propose WaterSplat-SLAM, achieving robust camera tracking and high-quality underwater scene modeling by RGB input. Our method extracts semantic medium information from images to suppress the adverse effects of water on camera tracking. We then incorporate semantic information into two-view 3D reconstruction prior to generating multi-view consistent depth maps and globally consistent camera poses, providing a high-quality initialization for the Gaussian map. To jointly model the underwater objects and the surrounding medium, we propose a semantic-guided rendering approach with an online medium-aware Gaussian map, avoiding misrepresenting the volumetric medium with Gaussian primitives. To ensure global consistency, we also adaptively adjust Gaussian primitives in response to loop closure and keyframe updates, and merge overlapping primitives across loop-closure frames to reduce redundancy. Across extensive experiments, it exceeds the NeRF-based and 3DGS-based SLAM methods in both tracking and mapping accuracy.

Our main contributions are as follows:
\begin{itemize}
    \item We propose a novel photorealistic monocular underwater SLAM system capable of robust pose estimation and photorealistic dense mapping;

    
    \item {We present an online medium-aware Gaussian map with a semantic-guided rendering, avoiding misrepresenting volumetric medium with Gaussian primitives;}
    

    \item {We design an adaptive management strategy that models the underwater environment in a compact manner, which significantly reduces the size of Gaussian primitives while maintaining rendering quality.}
\end{itemize}

\vspace*{-1.0mm}
\section{RELATED WORKS}
\vspace*{-1mm}
\subsection{{Underwater SLAM System}}
\vspace*{-1mm}
{
Recent research in sparse underwater SLAM and odometry focuses on enhancing robustness in feature-poor and visually degraded environments. SVIn2~\cite{SVIn2} and Vargas's work~\cite{vargas2021robust} enhance robustness by tightly coupling vision and IMU with acoustic sensors. In cases of total visual blackout, Joshi's work~\cite{joshi2023sm,joshi2023hybrid} combine kinematic characteristics of underwater robots to propagate attitude when the visual
inertial odometry (VIO) fails. DeepVL~\cite{singh2025deepvl} leverages recurrent neural networks to learn robot-centric velocity from motor commands and inertial cues, while Ferrera's work~\cite{ferrera2019real} focuses on a robust visual frontend specifically designed for turbid and dynamic environments.}

To tackle underwater online dense mapping and state estimation, many prior works have fused sonar, structured light, and laser sensing to obtain dense point clouds in visual SLAM systems. Wang's work~\cite{wang2020acoustic} couples an acoustic camera, generating 3D local maps via occupancy mapping and refining the global map via graph optimization to achieve dense underwater 3D mapping. 
Ou's work~\cite{ou2024structured} uses binocular structured light for collision-free navigation and visual dense mapping in unknown dark underwater environments. Hitchcox's work~\cite{hitchcox2023improving} and Hybrid-VINS~\cite{Hybrid-VINS} fuses laser and IMU with visual features for 3D dense underwater mapping.

However, these methods rely heavily on multi-sensor fusion, and their point-based or occupancy-based maps are limited for photorealistic rendering. In contrast, our approach uses only monocular RGB input and proposes an online medium-aware Gaussian map with semantic-guided rendering and adaptive management strategy, fully leveraging 3DGS for dense, photorealistic maps.
\vspace*{-3mm}
\begin{figure*}[htb]
    \centering
    \includegraphics[width=1.0\linewidth]{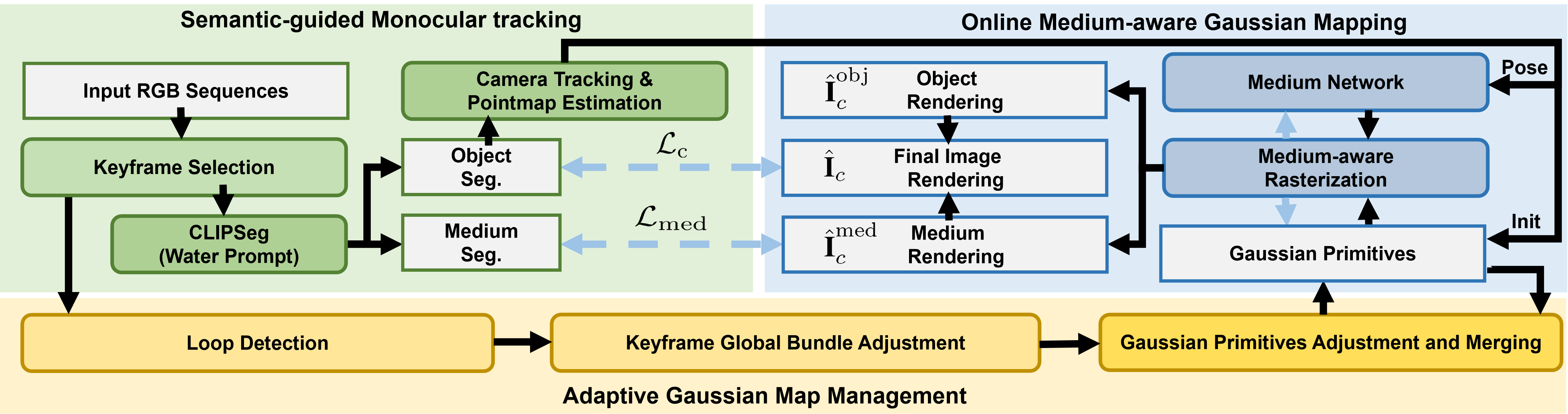}
    \caption{System overview of WaterSplat-SLAM: The system takes an RGB sequence as input and generates an online medium-aware Gaussian map. The RGB sequence is input into the segmentation model to generate semantic segmentation of images. {The object regions are then fed into the Camera Tracking and Pointmap Estimation module to generate keyframe poses.} Then, Gaussian primitives are initialized through keyframe poses and object parts. {For each encoded ray vector, the Medium Network predicts the medium parameters. The Gaussian map is optimized using a semantic-guided photometric loss. Upon loop closure, we also perform adaptive adjustment and merging of Gaussian primitives.}}
    \vspace*{-6mm}
    \label{fig: pipeline}
\end{figure*}
\subsection{Photorealistic Monocular Visual SLAM}
\vspace{-1mm}
MonoGS~\cite{Matsuki_2024_CVPR} is the first real-time 3DGS SLAM system for monocular input. To improve tracking efficiency, Photo-SLAM~\cite{huang2024photo} initializes Gaussian primitives with estimated ORB features and densifies the map via an image pyramid. Splat-SLAM~\cite{sandstrom2024splat} and GLORIE-SLAM~\cite{zhang2024glorie} introduce global bundle adjustment into Gaussian-based SLAM to enhance pose accuracy. Leveraging depth priors, HI-SLAM~\cite{zhang2023hi} uses joint depth and scale adjustment (JDSA) to resolve scale ambiguity. OpenGS-SLAM~\cite{yu2025rgb} employs DUSt3R~\cite{wang2024dust3r} to generate consistent multi-view pointmaps with differentiable camera poses. S3PO-GS~\cite{cheng2025outdoor} improves OpenGS-SLAM~\cite{yu2025rgb} with self-consistent 3DGS pointmap tracking and patch-based dynamic mapping via MASt3R~\cite{leroy2024grounding}. 

Different from previous works, to better adapt 3DGS representation in underwater scenarios, we design an online medium-aware Gaussian map to better adapt 3DGS to underwater scenarios by implicitly modeling the water medium. Furthermore, to better guide the medium network in learning water-specific properties, we leverage semantic masks to segment the water regions and design a semantic-guided rendering strategy, thereby preventing the water medium from being incorrectly represented by explicit primitives.

\vspace{-1mm}
\section{METHOD}  \label{sec: method}
\vspace*{-2mm}
\subsection{System Overview}
\vspace*{-1mm}

As illustrated in Fig.~\ref{fig: pipeline}, our WaterSplat-SLAM, from monocular input, enables robust underwater camera tracking and generates an online medium-aware Gaussian map with high-fidelity rendering.

The RGB sequence is segmented to obtain semantic masks. The object region (excluding water) is fed into the multi-view consistent Camera Tracking and Pointmap Estimation module to estimate keyframe poses and pointmaps, which then initialize and increment Gaussian primitives. Each encoded ray vector is passed through a Medium Network to predict three medium attributes: {$\boldsymbol{\sigma}^{\text{attn}}$ (attenuation density), $\boldsymbol{\sigma}^{\text{bs}}$ (backscatter density), and $\mathbf{c}^{\text{med}}$ (medium color)}. The Gaussian map is optimized in parallel using a semantic-guided photometric loss. Upon loop closure and global BA, Gaussian primitives are adaptively adjusted and merged to maintain global consistency, reducing overlap between associated keyframes.

{Different from WaterSplatting~\cite{Watersplatting} and SeaThru-NeRF~\cite{SeaThru_NeRF}, WaterSplat-SLAM proposes an online and incremental underwater SLAM system with photorealistic rendering, enhancing interactivity for real-time applications. Notably, we couple semantic medium information into online medium-aware Gaussian mapping, guiding the Medium Network to focus on learning inherent water properties and thereby improving the accuracy of medium attribute prediction. Meanwhile, the adaptive management of the Gaussian map maintains global consistency and reduces the size of Gaussian primitives while maintaining rendering quality, thereby significantly lowering the storage requirements during underwater robotic missions.}

\vspace*{-2mm}
\subsection{Semantic-guided Monocular Tracking} \label{sec:tracking and BA}
\vspace*{-1mm}

This module estimates multi-view consistent camera poses and pointmaps in underwater scenes from monocular input, providing high-quality structural initialization for the Gaussian map. To suppress the influence of water on camera tracking, we extract semantic medium information from images. We further incorporate semantic information into two-view 3D reconstruction prior to generating multi-view consistent depth maps and globally consistent camera poses.

We leverage MASt3R\cite{leroy2024grounding}, which takes a pair of images {$\mathbf{I}_i, \mathbf{I}_j \in \mathbb{R}^{H \times W \times 3}$} as input, and outputs pointmaps $\mathbf{X}_i^i, \mathbf{X}_i^j \in \mathbb{R}^{H \times W \times 3}$ with their features $\mathbf{D}_i^i, \mathbf{D}_i^j \in \mathbb{R}^{H \times W \times d}$ and confidences $\mathbf{Q}_i^i, \mathbf{Q}_i^j \in \mathbb{R}^{H \times W \times 1}$. {Here, $\mathbf{X}_i^i$ denotes the pointmap of image $i$ in its own camera coordinate frame, $\mathbf{X}_i^j$ denotes the pointmap of image $j$ represented in the coordinate frame of camera $i$. The same notation applies to the feature maps $\mathbf{D}_i^i, \mathbf{D}_i^j$ and confidence maps $\mathbf{Q}_i^i, \mathbf{Q}_i^j$.} While these confidences affect tracking performance as in ~\cite{murai2024mast3r}, we exclude water features from camera tracking. A text-guided image segmentation model provides prior information about water regions. Specifically, CLIPSeg~\cite{clipseg} is adopted to generate binary masks for water in images {$\mathbf{I}_i, \mathbf{I}_j$}:
\begin{align}
\begin{aligned}
\!
\mathbf{M}_i \! = \! \text{CLIPSeg}(\mathbf{I}_i, p_{water}),\,  \mathbf{M}_j \! = \! \text{CLIPSeg}(\mathbf{I}_j, p_{water}),
\label{watermask}
\end{aligned}
\end{align}
where $p_{water}$ denotes the text prompt for water, {$\mathbf{M}_i, \mathbf{M}_j \in \mathbb{R}^{H \times W \times 1}$} are the 2D masks of the image {$\mathbf{I}_i, \mathbf{I}_j$}. To eliminate the influence of water during camera tracking, we zero the confidence of water pixels and update as:
\vspace*{-1mm}
{
\begin{align}
\begin{aligned}
\mathbf{Q}_i^i &= \mathbf{Q}_i^i \odot (\mathbf{1} - \mathbf{M}_i), \quad
\mathbf{Q}_i^j = \mathbf{Q}_i^j \odot (\mathbf{1} - \mathbf{M}_j).
\label{water_mask}
\end{aligned}
\end{align}}
\vspace*{-5mm}

Since the system is a monocular modality with scale ambiguity, the camera pose is defined as:
\begin{align}
\mathbf{T} = \begin{bmatrix}
s\mathbf{R} & \mathbf{t} \\
\mathbf{0} & 1
\end{bmatrix} \in  \textbf{Sim}(3), \quad \mathbf{T}^{\text{SE3}} = \begin{bmatrix}
\mathbf{R} & \mathbf{t} \\
\mathbf{0} & 1
\end{bmatrix} \in  \textbf{SE}(3),
\label{T_sim3}
\end{align}
where $\mathbf{R} \in \text{SO}(3),\mathbf{t} \in \mathbb{R}^3$. Tracking is implemented as~\cite{murai2024mast3r}. 
With the last keyframe's pointmap estimation $\tilde{\mathbf{X}}_{k}^k$, current frame's pointmap estimation $\mathbf{X}_{f}^f$, and their matching relations $\mathbf{m}_{f,k}$, $\mathbf{T}_{kf}$ is obtained by minimizing the weighted points matching error:
\vspace{-1mm}
\begin{align}
\sum_{m,n \in \mathbf{m}_{f,k}} \left\| q_{m,n}(\tilde{\mathbf{X}}_{k,n}^k - \mathbf{T}_{kf} \mathbf{X}_{f,m}^f)\right\|_\rho,
\end{align}
\vspace*{-1.5mm}
where $\|\cdot\|_\rho$ denotes Huber norm, {
$q_{m,n} \! = \! \sqrt{\mathbf{Q}^f_{f,m} \mathbf{Q}^k_{f,n}}$, and $\mathbf{Q}^f_{f,m}$, $\mathbf{Q}^k_{f,n}$ are the matching confidence proposed in ~\cite{duisterhof2024mast3r}.}

\subsection{Online Medium-aware Gaussian Mapping}
\label{sec:3DGS_medium_illum}
\begin{figure}[t]
    \centering
    \includegraphics[width=0.49\textwidth]{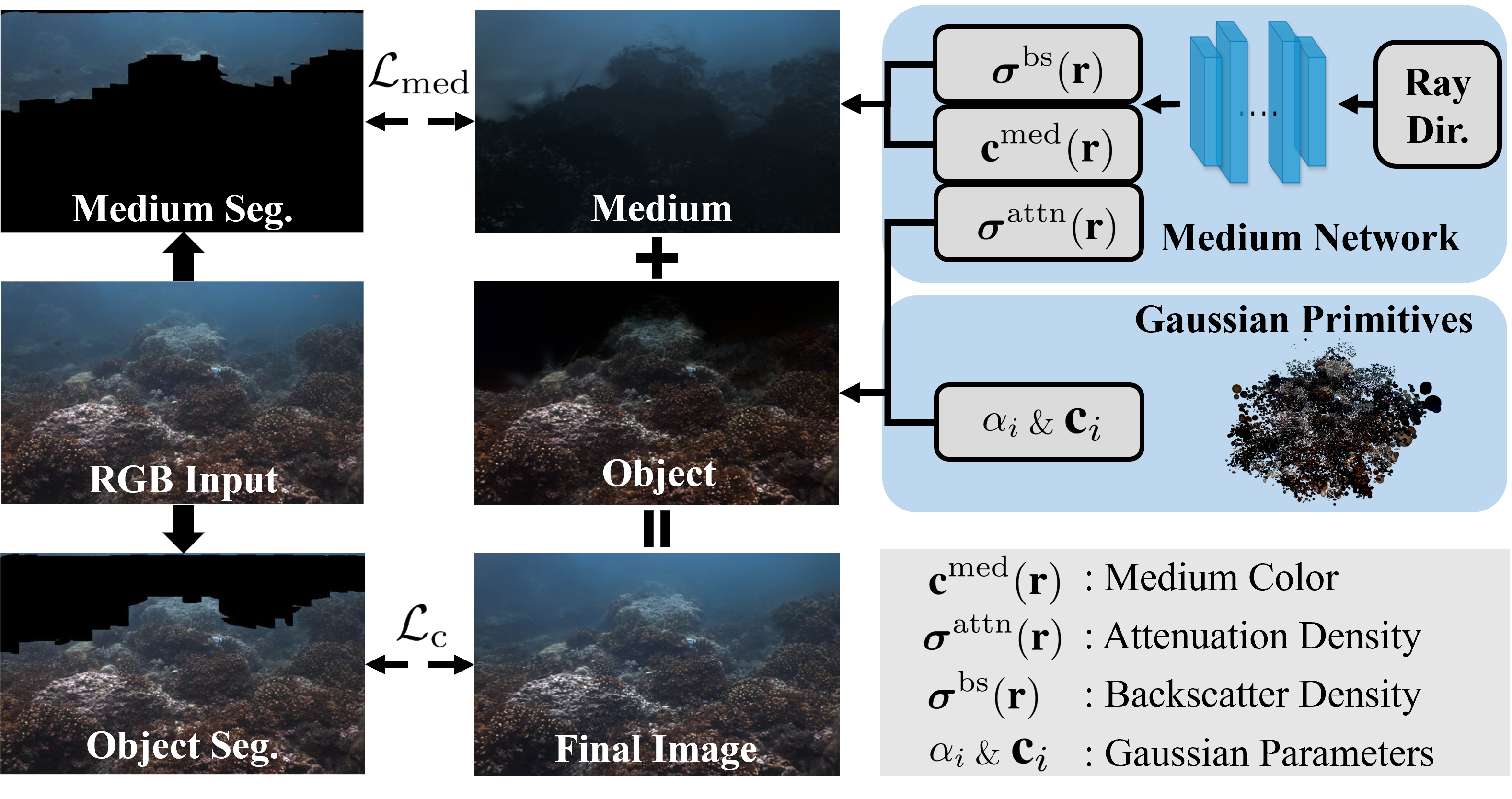}
    \vspace*{-5mm}
    \caption{Illustration of medium-aware Gaussian mapping: Encoded ray vectors are passed through the Medium Network to predict three medium attributes: 
    {$\sigma^{\text{attn}}$ (attenuation density), $\sigma^{\text{bs}}$ (backscatter density), and $c^{\text{med}}$ (medium color)}. During rendering, the contributions of the medium and objects are explicitly separated.}
    \vspace*{-4mm}
    \label{fig: gaussian map}
    \vspace*{-3.0mm}
\end{figure}

\vspace{-0.5mm}
This module generates an online and incremental medium-aware Gaussian map that precisely models scattering medium and geometric objects, enabling photorealistic mapping in underwater environments. As illustrated in Fig. \ref{fig: gaussian map}, the map incorporates three medium parameters through a Medium Network, and integrates them into the rasterization function. We leverage poses, pointmaps, and semantic information from Sec. \ref{sec:tracking and BA} for Gaussian primitives initialization and increment. We also design a semantic-guided photometric loss to guide the Medium Network to learn medium attributes.

3DGS is adopted as our map representation, where the scene is composed of a set of anisotropic Gaussian primitives $\mathcal{G}$. Each Gaussian primitive $\mathcal{G}^{i}$ contains five properties: color, scale, orientation, position and opacity, denoted as {{$\{\mathbf{c}_i, \mathbf{S}_{i},  \mathbf{R}_i, \bm{\mu}_i, \Lambda_i \}$}}. Given a 2D image plane and its camera pose $\mathbf{T}^{\mathrm{SE3}} = \begin{bmatrix}
\mathbf{R} & \mathbf{t} \\
\mathbf{0} & 1
\end{bmatrix} $, the 2D coordinate and covariance of each Gaussian primitive are obtained as:
\vspace*{-1mm}
\begin{align}
\hat{\bm{\mu}}_i=\pi((\mathbf{T}^{\text{SE3}})^{-1}\bm{\mu}_i), \quad \hat{\bm{\Sigma}}_i=\mathbf{J}\mathbf{R}^T\bm{\Sigma}_i\mathbf{R}\mathbf{J}^{T},
\end{align}
where $\pi$ is the projection operation and $\mathbf{J}$ is the Jacobian of linearized projective transformation, $\mathbf{\Sigma}_i$ is expressed as $\mathbf{\Sigma}_i = \mathbf{R}_i\mathbf{S}_{i}\mathbf{S}_{i}^T\mathbf{R}_i^T$. The opacity is computed as:
\vspace*{-2.0mm}
{
\begin{align}
\alpha_i = \Lambda_i \exp{(-\frac{1}{2} (\mathbf{p} - \hat{\bm{\mu}}_i)^T \hat{\bm{\Sigma}}_i^{-1} (\mathbf{p} -\hat{\bm{\mu}}_i))},
\end{align}}
{where $\mathbf{p}$ is the pixel coordinate}. After projecting Gaussian primitives and sorting by depth, {the object color of a pixel is computed via $\alpha$-blending of the Gaussian primitives},
\vspace{-1.5mm}
{
\begin{align}
\hat{\mathbf{C}}(\mathbf{r}) = \sum_{i=1}^{N} \alpha_i \mathbf{c}_i \prod_{j=1}^{i-1} (1 - \alpha_j) = \sum_{i=1}^{N} T_i \alpha_i \mathbf{c}_i,
\end{align}}
where $T_i = \prod_{j=1}^{i-1} (1 - \alpha_j)$ is the transmittance. Following~\cite{Watersplatting}, to model underwater image formation with attenuation and backscatter effects through medium, three parameters is added in $\alpha$-blending: {$\boldsymbol{\sigma}^{\text{attn}}$ (attenuation density), $\boldsymbol{\sigma}^{\text{bs}}$ (backscatter density), and $\mathbf{c}^{\text{med}}$ (medium color), each dependent on ray direction $\textbf{r}$}. These parameters are predicted by a Medium Network (MLP): {$\boldsymbol{\sigma}^{\mathrm{attn}}(\mathbf{r}), \boldsymbol{\sigma}^{\mathrm{bs}}(\mathbf{r}), \mathbf{c}^{\mathrm{med}}(\mathbf{r}) = \mathrm{MLP}(\mathrm{enc}(\mathbf{r}))$}, where $\mathrm{enc}$ is 16-dim spherical harmonics (SH) encoding. We rewrite our object rendering function as:
\vspace{-3.0mm}
\begin{align}
\hat{\mathbf{C}}^{\mathrm{obj}}(\mathbf{r}) = \sum_{i=1}^{N} T_i \alpha_i \exp(-\boldsymbol{\sigma}^{\mathrm{attn}}(\mathbf{r}) t_i) \mathbf{c}_i, 
\end{align}
and acquire medium rendering as:
\vspace{-1.3mm}
\begin{align}
\hat{\mathbf{C}}^{\mathrm{med}}(\mathbf{r}) = \mathbf{c}^{\mathrm{med}}(\mathbf{r}) 
\sum_{i=1}^{N} T_i \bigl[ 
    & \exp(-\boldsymbol{\sigma}^{\mathrm{bs}}(\mathbf{r}) t_{i-1}) \nonumber \\
  - & \exp(-\boldsymbol{\sigma}^{\mathrm{bs}}(\mathbf{r}) t_i) \bigr],
\label{medium_rendering}
\end{align}
where $t_i$ is the depth of the $i$-th Gaussian primitive with $t_0 = 0$, {$\boldsymbol{\sigma}^{\mathrm{attn}}(\mathbf{r})$, $\boldsymbol{\sigma}^{\mathrm{bs}}(\mathbf{r})$ and $\mathbf{c}^{\mathrm{med}}(\mathbf{r})$} are learned through the Medium Network whose input is ray direction {$\mathbf{r}$}. And the final rendered image is obtained as:

\vspace*{-4mm}
\begin{align}
{
\hat{\mathbf{C}}^{\mathrm{render}}(\mathbf{r}) = \hat{\mathbf{C}}^{\mathrm{obj}}(\mathbf{r}) + \hat{\mathbf{C}}^{\mathrm{med}}(\mathbf{r}).
}
\label{C_render}
\end{align}
Clear object rendering (without medium) is defined as:
\vspace*{-2mm}
{
\begin{align}
\hat{\mathbf{C}}^{\mathrm{clr}}(\mathbf{r}) = \sum_{i=1}^{N} \alpha_i \mathbf{c}_i \prod_{j=1}^{i-1} (1 - \alpha_j) = \sum_{i=1}^{N} T_i \alpha_i \mathbf{c}_i.
\label{clr}
\end{align}}

\vspace{-1mm}
For efficiency, we avoid adding Gaussian primitives in regions already sufficiently represented by existing Gaussian primitives.
Specifically, we compute the cumulative opacity at each pixel coordinate {$\mathbf{p}$: $\mathbf{A}(\mathbf{p}) =  \sum_{i=1}^{N} \alpha_i \prod_{j=1}^{i-1} \left(1 - \alpha_j\right)$}. {A pixel is labeled as incomplete if its cumulative opacity is lower than a threshold and it is not labeled as water by the segmentation mask.} The candidate set of incomplete points for adding new Gaussian primitives in keyframe $k$ is:
\vspace{-1.5mm}
{
\begin{align}
\mathbf{X}_k^{\mathrm{add}} = \mathbf{X}_k\odot(\mathbf{1}-\mathbf{M}_k)\odot(\mathbf{A}<\tau_A).
\end{align}}
New Gaussian primitives are then sampled from $\mathbf{X}_k^{\mathrm{add}}$ with a downsampling rate.

{We observe that optimizing Gaussian splatting using only a photometric loss between the rendered image and the ground truth tends to misattribute water-induced appearance effects to Gaussian primitives.} Thus, we design our loss functions to recover both photometric and geometric information in underwater scenes, while leveraging our semantic masks defined in Eq. \ref{watermask} to avoid misrepresenting the scattering medium with Gaussian primitives.

\textbf{Semantic-guided Photometric Loss:}  
For the rendered image {\( \hat{\mathbf{I}}_c  = \hat{\mathbf{I}}_c^{\mathrm{obj}} + \hat{\mathbf{I}}_c^{\mathrm{med}} \)} as in Eq. \ref{C_render} and the original image {\( {\mathbf{I}}_c \)}, we minimize the photometric loss between them. However, we hope that the pixels corresponding to the water medium should be expressed solely in {$\hat{\mathbf{I}}_c^{\mathrm{med}}$}. With the Water Mask {$\mathbf{M}_c$} for image {$\mathbf{I}_c$} (as in Eq. \ref{water_mask}), we design our medium photometric loss as:
\vspace{-1.5mm}
{
\begin{align}
\begin{aligned}
\mathcal{L}_\mathrm{med}
=  \left\lvert \bigl(\mathbf{M}_c\odot(\hat{\mathbf{I}}_c^{\mathrm{med}} - \mathbf{I}_c\bigr)\right\rvert_1, 
\end{aligned}
\end{align}}
and for the remaining pixels, the loss is:
\vspace{-1.5mm}
{
\begin{align}
\begin{aligned}
\mathcal{L}_c = \left\lvert \bigl(\mathbf{1}-\mathbf{M}_c)\odot(\hat{\mathbf{I}}_c  - \mathbf{I}_c\bigr)\right\rvert_1. 
\end{aligned}
\end{align}}
\vspace{-1.5mm}
The final semantic-guided photometric loss is:
\vspace{-0.3mm}
{
\begin{align}
\begin{aligned}
\mathcal{L}_\mathrm{sempho}
=& (1 - \lambda_\mathrm{ssim})(\mathcal{L}_\mathrm{med} + \mathcal{L}_c) \\ + &\lambda_\mathrm{ssim}(1-\mathrm{SSIM}(\hat{\mathbf{I}}_c, \mathbf{I}_c)),
\end{aligned}
\end{align}}
where the SSIM loss improves structural similarity.

\textbf{Geometric Loss:}  
Our geometric loss suppresses slender Gaussian primitives that may introduce artifacts. We apply isotropic loss on Gaussian primitives. Different from~\cite{Matsuki_2024_CVPR}, we only apply isotropic loss on those primitives contributing to image rendering to avoid scale distortion of other Gaussian primitives. We define our Gaussian visibility of current frame as $\text{VIS}_c$, then the isotropic loss is:

\vspace{-2mm}
\begin{align}
\mathcal{L}_s = \sum_{i \in \text{VIS}_c} |\mathbf{S}_{i} - \bar{\mathbf{S}}_i|,
\end{align}
{where $\bar{\mathbf{S}}_i$ is the mean of the three scale components of the $i$-th Gaussian primitive.} Finally, our total loss function between image {\(\hat{\mathbf{I}}_c \)} and image {\(\mathbf{I}_c\)}  is expressed as:
\begin{align}
\mathcal{L}(\hat{\mathbf{I}}_c, \mathbf{I}_c) = \lambda_\mathrm{sempho}\mathcal{L}_{\mathrm{sempho}} + 
\lambda_s\mathcal{L}_{s} .
\end{align}
\vspace*{-8mm}
\subsection{Adaptive Gaussian Map Management}
\begin{figure}[tb]
    \centering
    \includegraphics[width=1.0\linewidth]{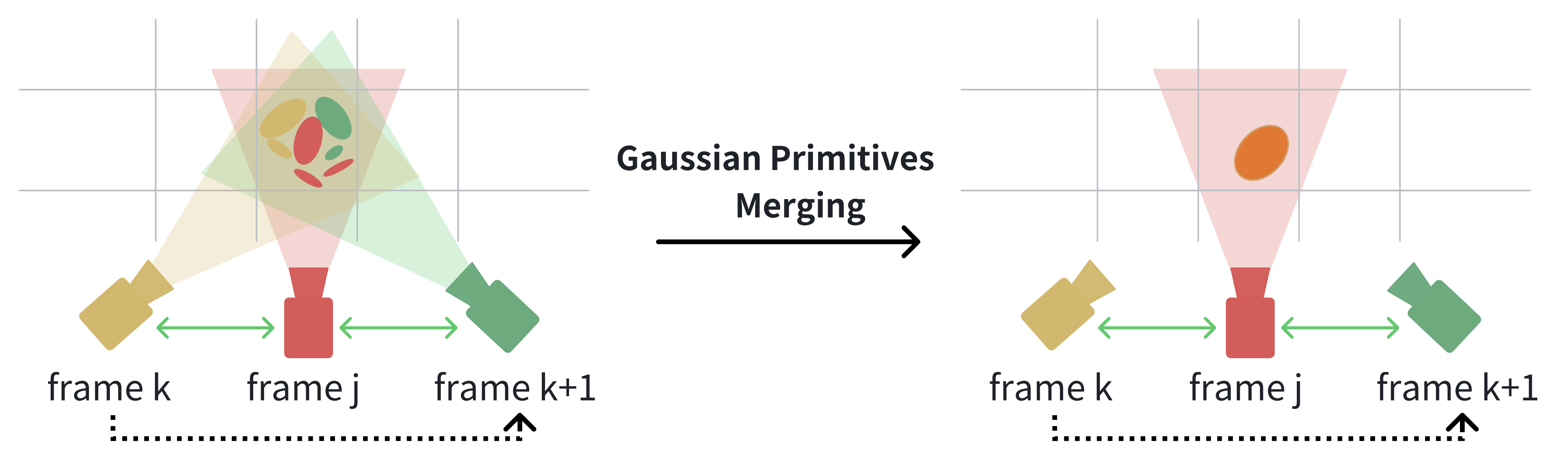}
    \vspace{-7mm}
    \caption{Gaussian primitives merging pipeline: When consecutive keyframes $k, k+1$ detect loop closure with history frame $j$, we extract their anchored Gaussian primitives and establish a 3D voxel grid. All primitives falling within the same voxel are then merged into a single Gaussian primitive.}
    \label{fig: merge pipeline}
    \vspace*{-7mm}
\end{figure}
\label{sec:densification_update}
This module maintains global consistency during loop closure and global BA, while reducing the size of Gaussian primitives bound to the current and loop closure frames. 
We implement backend optimization as~\cite{murai2024mast3r}, jointly minimizing projection errors for all edges $\mathcal{E}$ in the graph:
\begin{align}
\sum_{i,j \in \mathcal{E}} \sum_{m,n \in \mathbf{m}_{i,j}} \left\| q_{m,n}\left(\mathbf{p}_{i,m}^i - \pi \left( \mathbf{T}_{ij} \tilde{\mathbf{X}}_{j,n}^j \right) \right)\right\|_\rho,
\end{align}
{where $\mathbf{p}_{i,m}^i$ is the 2D pixel coordinate of the $m$-th feature in frame $i$, and $\tilde{\mathbf{X}}_{j,n}^j$ is the 3D world point associated with the $n$-th matched feature in frame $j$.} After the backend optimization process, keyframe poses are updated from $\mathbf{T}_k = \begin{bmatrix}
s_k\mathbf{R}_k & \mathbf{t}_k \\
\mathbf{0} & 1
\end{bmatrix}$ to $\mathbf{T}^{\prime}_k = \begin{bmatrix}
s_k^{\prime}\mathbf{R}_k^{\prime} & \mathbf{t}_k^{\prime} \\
\mathbf{0} & 1
\end{bmatrix}$. To ensure global consistency of the Gaussian map, we adjust the associated Gaussian primitives accordingly. Define keyframe $k$ as the anchored frame of the Gaussian primitive $\mathcal{G}^{i}$ if this primitive is added to global map through keyframe $k$. Assuming the properties of Gaussian primitive $\mathcal{G}^{i}$, $ \{\mathbf{S}_{i},  \mathbf{R}_i, \bm{\mu}_i\} $ are consistent in the coordinate of its anchored keyframe $k$, we update them as:
\vspace{-1mm}
\begin{align}
\begin{aligned}
&\mathbf{S}_{i}^{\prime} = \frac{s_k^{\prime}}{s_k}\mathbf{S}_{i},\,
\mathbf{R}_{i}^{\prime} = \mathbf{R}_{k}^{\prime}\mathbf{R}_{k}^{-1}\mathbf{R}_{i} ,\,
\bm{\mu}_i^{\prime} = \mathbf{T}^{\prime}_k (\mathbf{T}_k)^{-1}
\bm{\mu}_i.
\end{aligned}
\end{align}

\vspace{-2.0mm}
This transformation preserves the geometric relationships among Gaussian primitives while aligning them with the refined keyframe poses, thereby maintaining the accuracy and completeness of 3D reconstruction.
\begin{figure*}[h!]
    \centering
    \includegraphics[width=1.0\linewidth]{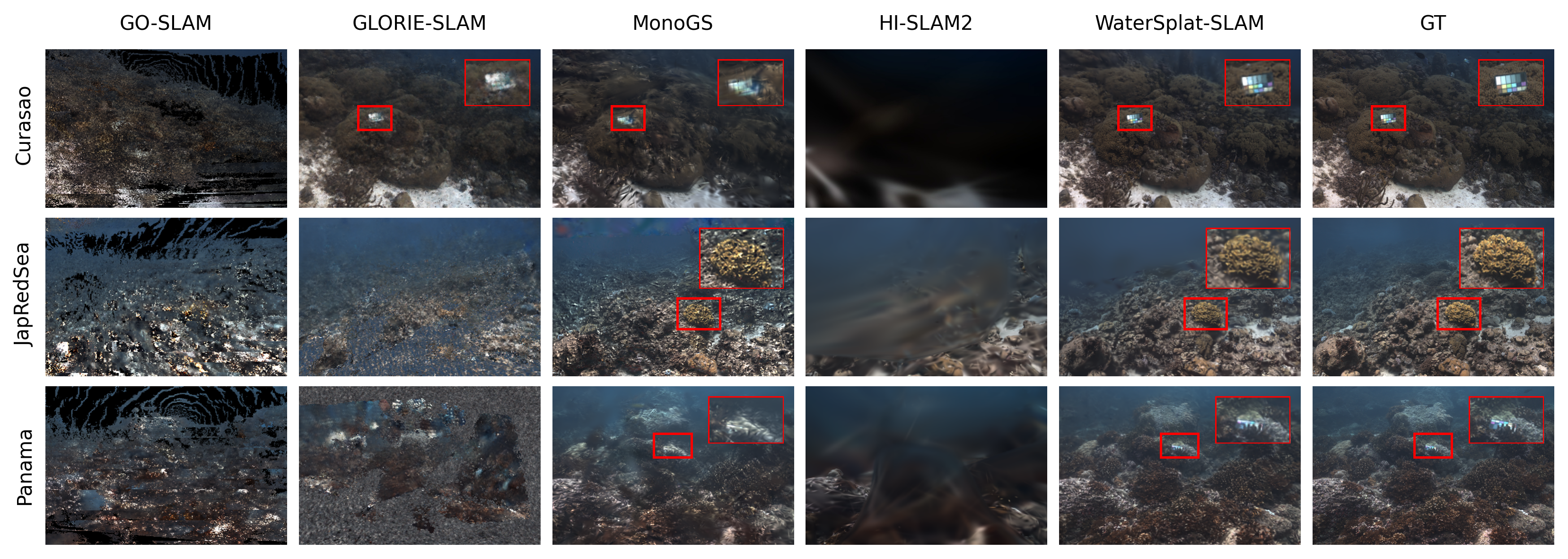}
    \vspace{-7mm}
    \caption{Detailed comparison of reconstruction results for Curacao, JapRedSea, and Panama sequences on SeaThru-NeRF dataset. {All three sequences exhibit the common challenge of turbid water and significant obscureness.} We select a non-keyframe which is not included in training views across all methods in three sequences. Specific regions are zoomed in to highlight reconstruction details. WaterSplat-SLAM shows high-fidelity reconstruction and details, {demonstrating its strong capability in clearly rendering both foreground and background objects despite the challenging underwater conditions.}}
    \vspace{-2mm}
    \label{fig: seathru_comparison}
\end{figure*}

\begin{table*}[h!]
\centering
\caption{Rendering quality evaluations on SeaThru-NeRF dataset.}
\renewcommand{\arraystretch}{1.3}
\label{tab:seathru_render}
\setlength{\tabcolsep}{1.0pt}
\resizebox{1.0\textwidth}{!}{
\begin{tabular}{lcc
                *{3}{c}
                *{3}{c}
                *{3}{c}
                *{3}{c}
                *{3}{c}}
\toprule
\multirow{2}{*}{Method} & \multirow{2}{*}{Type} & 
\multicolumn{3}{c}{Curasao} & 
\multicolumn{3}{c}{RedSea} & 
\multicolumn{3}{c}{JapRedSea} & 
\multicolumn{3}{c}{Panama} & 
\multicolumn{3}{c}{Avg} \\
\cmidrule(lr){3-5} \cmidrule(lr){6-8} \cmidrule(lr){9-11} \cmidrule(lr){12-14} \cmidrule(lr){15-17}
& & PSNR$\uparrow$ & SSIM$\uparrow$ & LPIPS$\downarrow$ & 
PSNR$\uparrow$ & SSIM$\uparrow$ & LPIPS$\downarrow$ & 
PSNR$\uparrow$ & SSIM$\uparrow$ & LPIPS$\downarrow$ & 
PSNR$\uparrow$ & SSIM$\uparrow$ & LPIPS$\downarrow$ & 
PSNR$\uparrow$ & SSIM$\uparrow$ & LPIPS$\downarrow$ \\
\midrule
GO-SLAM\cite{Zhang_2023_ICCV} &  \multirow{2}*{NeRF} & 
18.70 & 0.353 & 0.795 & 
11.89 & 0.131 & 0.791 & 
15.72 & 0.294 & 0.841 & 
16.63 & 0.289 & 0.898 & 
15.74 & 0.267 & 0.831 \\

GLORIE-SLAM\cite{zhang2024glorie} &  & 
21.79 & 0.567 & 0.558 & 
16.02 & 0.415 & 0.575 & 
16.11 & 0.446 & 0.494 & 
18.79 & 0.452 & 0.523 & 
18.18 & 0.470 & 0.538 \\
\midrule
MonoGS\cite{Matsuki_2024_CVPR} &  \multirow{5}*{3DGS} & 
20.83 & 0.613 & 0.643 & 
Fail & Fail & Fail & 
14.76 & 0.465 & 0.553 & 
18.54 & 0.597 & 0.557 & 
18.04 & 0.558 & 0.584 \\

HI-SLAM2\cite{zhang2024hi} &  & 
15.24 & 0.247 & 0.913 & 
14.28 & 0.503 & 0.688 & 
12.04 & 0.277 & 0.782 & 
12.10 & 0.152 & 0.936 & 
13.42 & 0.295 & 0.830 \\

OpenGS-SLAM\cite{yu2025rgb} &  & 
Fail & Fail & Fail & 
Fail & Fail & Fail & 
17.11 & 0.573 & 0.581 & 
19.59 & 0.652 & 0.710 & 
18.35 & 0.613 & 0.646\\

S3PO-GS\cite{cheng2025outdoor} &  & 
20.18 & 0.604 & 0.612 & 
\textbf{20.01} & \textbf{0.621} & 0.445 & 
17.55 & 0.550 & 0.470 & 
19.51 & 0.595 & 0.485 & 
19.06 & 0.593 & 0.503 \\

\textbf{WaterSplat-SLAM} &  & 
\textbf{28.91} & \textbf{0.780} & \textbf{0.328} & 
18.01 & 0.521 & \textbf{0.441} & 
\textbf{23.68} & \textbf{0.761} & \textbf{0.301} & 
\textbf{25.95} & \textbf{0.723} & \textbf{0.270} & 
\textbf{24.14} & \textbf{0.696} & \textbf{0.335} \\
\bottomrule
\end{tabular}}
\vspace{-7mm}
\end{table*}
The relative pose between $\mathbf{T}_k^{\mathrm{SE3}}$ and $\mathbf{T}^{\prime\mathrm{SE3}}_k$  is characterized by a rotation angle \(\Delta \theta\) and a translational distance \(\Delta d\), which are derived from the delta transformation matrix \(\Delta \mathbf{T}^{\mathrm{SE3}} = (\mathbf{T}_k^{\mathrm{SE3}})^{-1}\mathbf{T}^{\prime\mathrm{SE3}}_k\). To maintain the rendering quality of the Gaussian map, we employ a keyframe-based validation mechanism. A keyframe is marked for re-rendering if its pose change exceeds predefined thresholds-i.e., \(\Delta d > d_{\mathrm{thresh}}\) or \(\Delta \theta > \theta_{\mathrm{thresh}}\).
Moreover, if the proportion of keyframes marked for re-rendering exceeds a certain threshold (e.g., \( \frac{\mathrm{\# re-render keyframes}}{\mathrm{\# total keyframes}} > GLOMAP_{\mathrm{thresh}} \)), it indicates significant drift or deformation in the Gaussian map. In such cases, a global mapping process is triggered, where all keyframes are re-rendered to ensure global consistency. 

We observe that when loop closure is detected through~\cite{tolias2020learning, duisterhof2024mast3r}, spatial overlap exists between Gaussian primitives bound to the current frame and the loop closure frame after Gaussian primitives adjustment. {To reduce the map storage and facilitate system deployment}, we further reduce redundancy by merging Gaussian primitives through voxel downsampling. As illustrated in Fig. \ref{fig: merge pipeline}, for each detected loop closure between a current frame and a loop closure frame, we extract all Gaussian primitives bound to either frame and voxelize them in the world coordinate system with voxel size \( v \). Within each non-empty voxel \( \mathcal{V} \), we merge all Gaussian primitives into a single representative primitive. The merging process computes the average of their attributes, including {color \( \mathbf{c} \)}, scale \( \mathbf{S} \), orientation \( \mathbf{R} \), position \( \bm{\mu} \), and opacity \( \Lambda \). The new Gaussian primitive is then bound to a randomly selected previous frame in \( \mathcal{V} \).

\subsection{Implementation Details}
\vspace{-1mm}
For two-view geometry prediction with MASt3R~\cite{leroy2024grounding} and Water Mask prediction, we employ pretrained models from [20] and [24], respectively. During SLAM
 process, we implement a parallel process for underwater Gaussian mapping on keyframes, with the proposed loss function. The loss weights are consistent across all experiments, with $\lambda_\mathrm{ssim} = 0.2$, $\lambda_\mathrm{sempho} = 1$ and $\lambda_{s} = 3$. The Medium Network (MLP) for acquiring medium parameters {$\boldsymbol{\sigma}^{\mathrm{attn}}$, $\boldsymbol{\sigma}^{\mathrm{bs}}$, and $\mathbf{c}^{\mathrm{med}}$} is a 4-layer network with 128 hidden units per layer. For mapping, each newly added keyframe is rendered for 50 iterations, and 5 randomly selected keyframes for 40 iterations. The opacity threshold $\tau_A$ for adding a new Gaussian is set to $0.9$. We set \(\  d_{\mathrm{thresh}} = 0.15\) and \(\theta_{\mathrm{thresh}} = 7.0\) for re-rendering keyframe judgement, and $GLOMAP_{\mathrm{thresh}} = 0.3$ for triggering global mapping. The voxel size \( v \) for Gaussian primitives merging is set to 0.35.

\vspace{-2mm}
\section{EXPERIMENT}
\vspace{-0.5mm}
We validate our WaterSplat-SLAM on two datasets: the open-source SeaThru-NeRF dataset~\cite{SeaThru_NeRF} and WaterSplat-SLAM dataset we collected with varying visibility conditions and geometric complexity. We firstly present the datasets and evaluation metrics, followed by qualitative comparisons of rendering quality, camera-tracking accuracy against recent methods. Finally, we conduct ablation studies to analyze the impact of different design choices.
\vspace{-1mm}
\begin{figure*}[ht]
    \centering
    \includegraphics[width=1.0\linewidth]{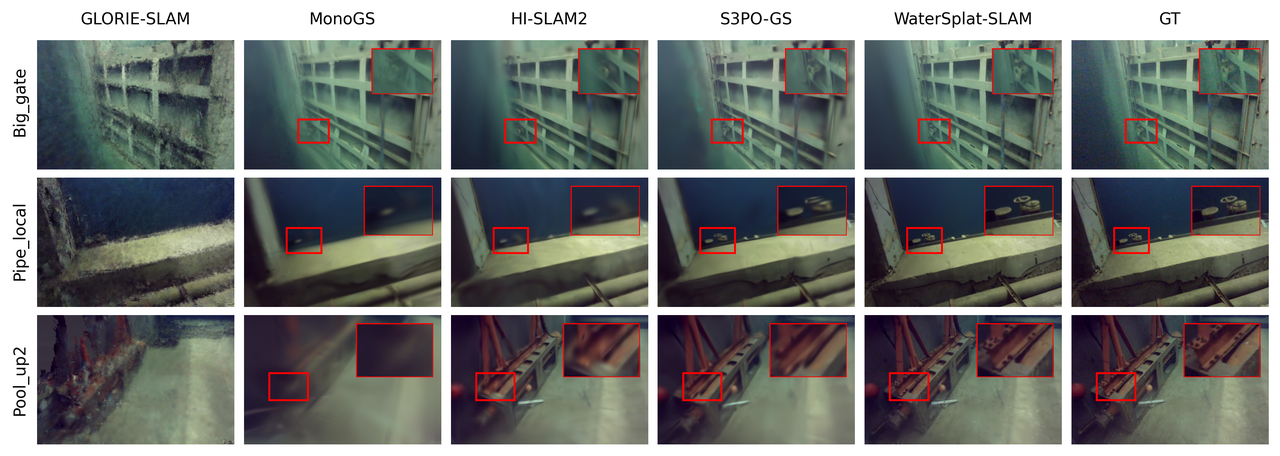}
    \vspace*{-5mm}
    \caption{Detailed reconstruction comparisons for Big\_gate and Pipe\_local and Pool\_up2 sequences on our dataset. We select a non-keyframe which is not included in training views for all methods across three sequences. Specific regions are zoomed in to highlight reconstruction details.}
    \vspace*{-3mm}
    \label{fig: ours_comparison}
\end{figure*}

\begin{table*}[h]
\centering
\vspace{-0mm}
\caption{Rendering quality evaluations on WaterSplat-SLAM dataset.}
\renewcommand{\arraystretch}{1.3}
\label{tab:tank_render}
\setlength{\tabcolsep}{1.0pt}
\resizebox{1.0 \textwidth}{!}{
\begin{tabular}{lcc
                *{3}{c}
                *{3}{c}
                *{3}{c}
                *{3}{c}
                *{3}{c}
                *{3}{c}}
\toprule
\multirow{2}{*}{Method} & \multirow{2}{*}{Type} & 
\multicolumn{3}{c}{Big\_gate} & 
\multicolumn{3}{c}{Pipe\_local} & 
\multicolumn{3}{c}{Pool\_up} & 
\multicolumn{3}{c}{Pool\_up2} & 
\multicolumn{3}{c}{Pool\_loop} & 
\multicolumn{3}{c}{Avg} \\
\cmidrule(lr){3-5} \cmidrule(lr){6-8} \cmidrule(lr){9-11} \cmidrule(lr){12-14} \cmidrule(lr){15-17} \cmidrule(lr){18-20}
& & PSNR$\uparrow$ & SSIM$\uparrow$ & LPIPS$\downarrow$ & 
PSNR$\uparrow$ & SSIM$\uparrow$ & LPIPS$\downarrow$ & 
PSNR$\uparrow$ & SSIM$\uparrow$ & LPIPS$\downarrow$ & 
PSNR$\uparrow$ & SSIM$\uparrow$ & LPIPS$\downarrow$ & 
PSNR$\uparrow$ & SSIM$\uparrow$ & LPIPS$\downarrow$ & 
PSNR$\uparrow$ & SSIM$\uparrow$ & LPIPS$\downarrow$ \\
\midrule
GO-SLAM\cite{Zhang_2023_ICCV} & \multirow{2}*{NeRF} & 
17.76 & 0.348 & 0.616 & 
14.20 & 0.269 & 0.743 & 
10.97 & 0.144 & 0.824 & 
19.83 & 0.510 & 0.609 & 
10.02 & 0.192 & 0.779 & 
14.56 & 0.293 & 0.714 \\

GLORIE-SLAM\cite{zhang2024glorie} &  & 
19.51 & 0.643 & 0.556 & 
21.54 & 0.682 & 0.597 & 
Fail & Fail & Fail & 
20.48 & 0.649 & 0.519 & 
Fail & Fail & Fail & 
20.51 & 0.658 & 0.557 \\
\midrule
MonoGS\cite{Matsuki_2024_CVPR} & \multirow{5}*{3DGS} & 
25.97 & \textbf{0.765} & 0.672 & 
23.71 & 0.860 & 0.649 & 
23.21 & 0.864 & 0.629 & 
26.44 & 0.897 & 0.572 & 
17.25 & 0.747 & 0.719 & 
23.32 & 0.827 & 0.648 \\

HI-SLAM2\cite{zhang2024hi} &  & 
24.01 & 0.650 & 0.575 & 
25.61 & 0.820 & 0.451 & 
21.39 & 0.762 & 0.591 & 
26.09 & 0.834 & 0.460 & 
22.41 & 0.746 & 0.618 & 
23.90 & 0.762 & 0.539 \\

OpenGS-SLAM\cite{yu2025rgb} &  & 
Fail & Fail & Fail & 
Fail & Fail & Fail & 
23.01 & 0.752 & 0.585 & 
24.76 & 0.880 & 0.570 & 
Fail & Fail & Fail &
23.89 & 0.816 & 0.578 \\

S3PO-GS\cite{cheng2025outdoor} &  & 
25.97 & 0.643 & 0.547 & 
27.90 & \textbf{0.885} & 0.593 & 
25.84 & 0.732 & 0.515 & 
27.02 & 0.896 & 0.553 & 
Fail & Fail & Fail &
26.68 & 0.789 & 0.552 \\

\textbf{WaterSplat-SLAM} &  & 
\textbf{29.55} & 0.748 & \textbf{0.324} & 
\textbf{30.20} & 0.861 & \textbf{0.238} & 
30.29 & 0.861 & \textbf{0.239} & 
\textbf{33.22} & 0.897 & \textbf{0.216} & 
\textbf{27.69} & 0.808 & \textbf{0.326} & 
\textbf{30.19} & 0.835 & \textbf{0.269} \\
\bottomrule
\end{tabular}}
\vspace{-5mm}
\end{table*}

\subsection{Datasets and Evaluation Metrics}
We evaluate our method on the SeaThru-NeRF dataset~\cite{SeaThru_NeRF}, which contains underwater images from four regions: Curasao, RedSea, JapRedSea and Panama, where light attenuation and backscatter effect are prominent. To further validate WaterSplat-SLAM's camera tracking and reconstruction performance on long sequences with loop closure, we collect the "WaterSplat-SLAM" dataset, by a pre-calibrated monocular camera in a controlled pool environment (approximately \(20\,\mathrm{m} \times 10\,\mathrm{m}\)). The dataset consists of six sequences: Big\_gate, Pipe\_local, Pool\_up, Pool\_up2, Pool\_loop, each representing different underwater scenarios. {Notably, both datasets feature turbid water with significant obscureness, resulting in clear near-range objects and blurred distant ones.} Ground-truth camera poses are reconstructed via SFM for quantitative evaluation. 

For performance evaluation, we assess system performance across three critical dimensions. For rendering quality, we employ three established metrics on novel view synthesis (NVS) assessment: PSNR (Peak Signal-to-Noise Ratio), SSIM (Structural Similarity Index), and LPIPS (Learned Perceptual Image Patch Similarity). We evaluate camera tracking accuracy using RMSE of Absolute Trajectory Error (ATE) on keyframes against ground-truth poses. All evaluations include comprehensive benchmarking against state-of-the-art monocular NeRF-SLAM and 3DGS-SLAM systems. Additionally, for rendering quality, we provide a comparison with the offline method WaterSplatting~\cite{Watersplatting}, where ground-truth poses are provided as input, and its training/test splits and iteration counts are aligned with our online system. For ATE evaluation, we additionally compare with ORB-SLAM3~\cite{campos2021orb} using RGB input. The best results are highlighted in bold.
\vspace{-1mm}

\subsection{Rendering Quality}
We quantitatively evaluate NVS performance by rendering on non-keyframes (excluded from training views) and comparing against state-of-the-art monocular SLAM systems: NeRF-based approaches (GO-SLAM~\cite{Zhang_2023_ICCV}, GLORIE-SLAM~\cite{zhang2024glorie}) and Gaussian splatting methods (MonoGS~\cite{Matsuki_2024_CVPR}, HI-SLAM2~\cite{zhang2024hi}, OpenGS-SLAM~\cite{yu2025rgb}, S3PO-GS~\cite{cheng2025outdoor}). Experiments are conducted on all four sequences of the SeaThru-NeRF dataset~\cite{SeaThru_NeRF} and our five custom tank sequences, using RGB-only input. 

As shown in Table. \ref{tab:seathru_render} and Fig. \ref{fig: seathru_comparison}, WaterSplat-SLAM achieves superior rendering quality across all SeaThru-NeRF sequences and metrics. Among the compared methods, HI-SLAM2~\cite{zhang2024hi} suffers from inaccurate normal and depth estimations in aquatic environments, resulting in suboptimal reconstructions. MonoGS~\cite{Matsuki_2024_CVPR}, despite its lack of depth priors, still delivers competitive rendering results on several sequences. S3PO-GS~\cite{cheng2025outdoor} demonstrates robust performance across all datasets, achieving particularly strong results on the challenging RedSea sequence. However, conventional 3DGS approaches exhibit certain limitations in modeling water scenes due to their direct application of Gaussian primitives without medium modeling. In contrast, our WaterSplat-SLAM, with its dedicated medium rendering and semantic-guided pipeline, effectively addresses these challenges and yields better performance in rendering quality with fewer artifacts.

Table. \ref{tab:tank_render} further demonstrates the robustness of our system in large-scale underwater scenes. While GO-SLAM~\cite{Zhang_2023_ICCV} suffers from limited rendering quality in these complex environments, methods such as HI-SLAM2~\cite{zhang2024hi} and MonoGS~\cite{Matsuki_2024_CVPR} exhibit more stable performance in structured areas of the tank scenes. S3PO-GS~\cite{cheng2025outdoor} also achieves competitive average results across the sequences. However, these approaches still face challenges in consistently modeling both the aquatic medium and scene objects. WaterSplat-SLAM maintains both accurate geometry and photorealistic outputs, achieving the best average PSNR, SSIM, and LPIPS metrics. Moreover, thanks to our Gaussian map's rapid adjustment and merging strategy, WaterSplat-SLAM achieves particularly high rendering quality in large-scale scenes with loop closures, as demonstrated in the Pool\_loop sequence.
\begin{table}[t]
\centering
\caption{PSNR comparison between WaterSplatting and WaterSplat-SLAM on WaterSplat-SLAM dataset.}\label{tab:offline_psnr_comparison}
\vspace{-2.0mm}
\setlength{\tabcolsep}{1.0pt}
\resizebox{0.473\textwidth}{!}{
\begin{tabular}{lcccccc}
\toprule
Method & Big\_gate & Pipe\_local & Pool\_up & Pool\_up2 & Pool\_loop & Avg \\
\midrule
{WaterSplatting~\cite{Watersplatting}} & {27.49} & {28.37} & {\textbf{30.62}} & {31.74} & {26.16} & {28.88} \\
\textbf{WaterSplat-SLAM} & \textbf{29.55} & \textbf{30.20} & 30.29 & \textbf{33.22} & \textbf{27.69} & \textbf{30.19} \\
\bottomrule
\end{tabular}}
\vspace{-2mm}
\end{table}

{
Table \ref{tab:offline_psnr_comparison} provides a comparison of the PSNR performance between WaterSplat-SLAM and the offline WaterSplatting~\cite{Watersplatting} across WaterSplat-SLAM datasets. Although our method operates in an online manner, it achieves better performance than WaterSplatting. This performance gain can be attributed primarily to the proposed semantic-guided rendering strategy in our methods. Our method achieves high-quality rendering while maintaining real-time capability.}
\vspace{-1.0mm}
\subsection{Camera Tracking Accuracy}

\begin{table}[t]
\centering
\caption{Tracking accuracy (ATE(m) $\downarrow$) on WaterSplat-SLAM dataset. }\label{tab:tank_tracking}
\vspace{-2mm}
\setlength{\tabcolsep}{1.0pt}
\resizebox{0.48\textwidth}{!}{
\begin{tabular}{lccccccc}
\toprule
Method & Big\_gate & Pipe\_local & Pool\_up & Pool\_up2 & Pool\_loop & Avg \\
\midrule
ORB-SLAM3 \cite{campos2021orb} & 0.091 & 2.768 & Fail & Fail & Fail & $\times$ \\
GO-SLAM \cite{Zhang_2023_ICCV} & {0.041} & 0.292 & {2.532} & {0.205} & {3.231} & 1.260 \\
GLORIE-SLAM \cite{zhang2024glorie} & {0.036} & {0.236} & Fail & \textbf{0.086} & Fail & $\times$ \\
MonoGS \cite{Matsuki_2024_CVPR}& 1.729 & 3.472 & 2.899 & 2.578 & 3.240 & 2.784 \\
HI-SLAM2 \cite{zhang2024hi} & \textbf{0.034} & \textbf{0.135} & {2.533} & 1.213 & {2.343} & 1.252 \\
OpenGS-SLAM\cite{yu2025rgb} & Fail & Fail & 2.063 & 2.249 & Fail & $\times$ \\
S3PO-GS\cite{cheng2025outdoor} & 0.114 & 0.592 & 1.593 & 0.834 & Fail & $\times$ \\
\textbf{WaterSplat-SLAM} & 0.066 & {0.164} & \textbf{0.289} & {0.224} & \textbf{0.353} & \textbf{0.219} \\
\bottomrule
\vspace*{-10mm}
\end{tabular}}
\end{table}
\vspace*{-1.0mm}
We quantitatively evaluate camera tracking accuracy by aligning trajectories to COLMAP SFM reconstructions on our underwater sequences
As shown in Table. \ref{tab:tank_tracking}, ORB-SLAM3~\cite{campos2021orb} fails catastrophically in three sequences due to severe feature degradation underwater, and GLORIE-SLAM~\cite{zhang2024glorie} fails in two sequences. MonoGS~\cite{Matsuki_2024_CVPR} suffers from the lack of depth priors, leading to geometrically inconsistent maps. Methods like HI-SLAM2~\cite{zhang2024hi} achieve competitive tracking accuracy in smaller or more structured scenes, such as Big\_gate and Pipe\_local, but exhibit degraded performance in larger-scale or more heavily water-dominated environments due to challenges in depth and normal estimation. Similarly, S3PO-GS~\cite{cheng2025outdoor} demonstrates strength in localized settings but shows limitations in maintaining global consistency during loop closures in expansive aquatic scenes. OpenGS-SLAM~\cite{yu2025rgb} fails on multiple sequences due to its lack of control over Gaussian primitive splitting. WaterSplat-SLAM maintains robust tracking across all sequences without failure. The exclusion of water features during tracking ensures our system's robustness and generalization across various underwater scenes.

\vspace{-2mm}
\subsection{Ablation Study}
\vspace{-1mm}
{We propose several key components: a semantic-guided water tracking and rendering method, and a Gaussian primitive adjustment and merging strategy.} We conduct our ablation study on each module of the four modules in Panama sequence and our dataset.

\begin{figure}[t!]
\centering
\includegraphics[width=0.492\textwidth]{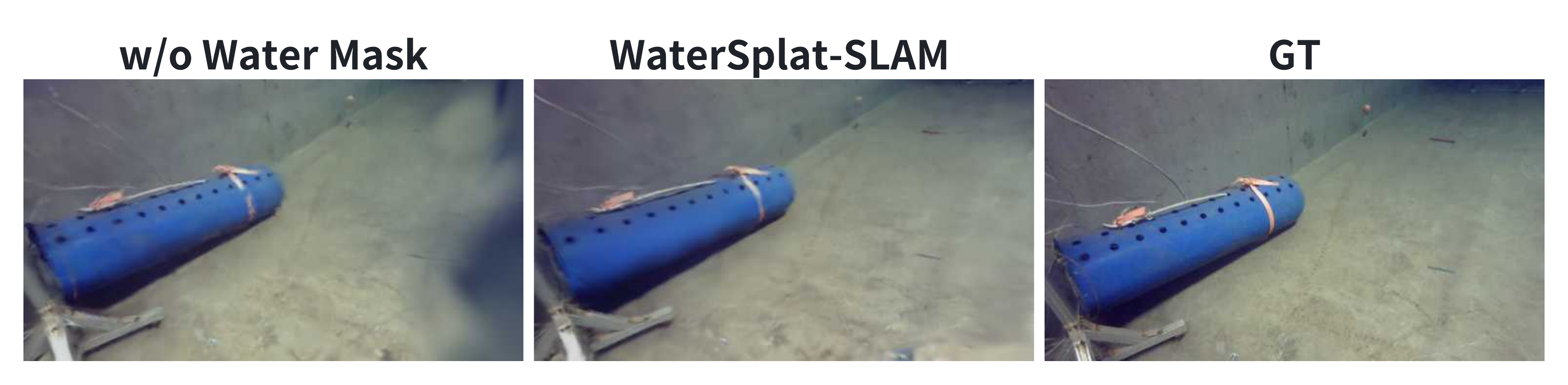}
\vspace*{-8mm}
\caption{NVS result w/o Water Mask module.}\label{fig:ablation pic}
\vspace*{-8mm}
\end{figure}

For the Panama scene, as in Table. \ref{tab:ablation_render}, our semantic-guided rendering module contributes to the system performance crucially. For Pool\_loop scene, as in Table. \ref{tab:ablation_render}, our primitives adjustment module also contributes to the system performance crucially, as it includes loop closure. As shown in Fig. \ref{fig:ablation pic}, removing of semantic-guided rendering module leads to incorrect reconstruction of medium and artifacts in novel view synthesis. As shown in Table. \ref{tab:ate_comparison}, the absence of the Water Mask degrades tracking accuracy in most scenarios. The slight increase of ATE in Pipe\_local with Water Mask is caused by incorrect semantic segmentation of water.

\begin{table}[t]
\centering
\caption{Ablation study of different modules in \textit{Panama} and \textit{Pool\_loop} on rendering performance}
\label{tab:ablation_render} 
\setlength{\tabcolsep}{2.0pt}
\resizebox{0.473\textwidth}{!}{
\begin{tabular}{lccccc}
\toprule
Dataset & Method & PSNR $\uparrow$ & SSIM $\uparrow$ & LPIPS $\downarrow$ \\
\midrule
\multirow{3}{*}{Panama} 
    & w/o Water Mask             &  24.85 &  0.701 &  0.357 \\
    & w/o Primitives Adjustment  &  25.63 &  0.705 &  0.326 \\
    & \textbf{WaterSplat-SLAM}   &  \textbf{25.95} & \textbf{ 0.723} &  \textbf{0.270} \\
\midrule
\multirow{3}{*}{Pool\_loop}
    & w/o Water Mask             &  27.21 &  0.797 &  0.353 \\
    & w/o Primitives Adjustment  &  26.56 &  0.790 &  0.376 \\
    & \textbf{WaterSplat-SLAM}   &  \textbf{27.69} &  \textbf{0.808} &  \textbf{0.326} \\
\bottomrule\end{tabular}}
\vspace{-2mm}
\end{table}

\begin{table}[t]
\vspace{-1mm}
\centering
\caption{Water Mask module in WaterSplat-SLAM dataset on ATE(m) 
$\downarrow$}
\label{tab:ate_comparison}
\vspace{-3mm}
\setlength{\tabcolsep}{1.5pt}
\resizebox{0.473\textwidth}{!}{
\begin{tabular}{lcccccc}
    \toprule
      & Big\_gate & Pipe\_local & Pool\_up & Pool\_up2 & Pool\_loop & Avg \\
    \midrule
    w/o Water mask & 0.077 & \textbf{0.160} &  0.323 &  0.274 &  0.375 & 0.242 \\
    \textbf{WaterSplat-SLAM} & \textbf{0.066} & 0.164 &  \textbf{0.289} &  \textbf{0.224} & \textbf{0.353} & \textbf{0.219} \\
    \bottomrule
\end{tabular}}
\vspace{-4mm}
\end{table}

\begin{table}[h]
\centering
\caption{Ablation study on primitives merge strategy in \textit{Pool\_loop}}
\label{tab:ablation_primitives}
\setlength{\tabcolsep}{1.5pt}
\resizebox{0.473\textwidth}{!}{
\begin{tabular}{lcccc}
    \toprule
    \multirow{2}{*}{Method} & \multirow{2}{*}{\makecell{Primitives \\ Size}} & \multicolumn{3}{c}{Reconstruction Quality} \\
    \cmidrule{3-5}
    & & \makecell{PSNR $\uparrow$} & \makecell{SSIM $\uparrow$} & \makecell{LPIPS $\downarrow$} \\
    \midrule
    w/o Primitives Merge & 137657 & 27.59 & 0.807 & \textbf{0.303} \\
    \textbf{WaterSplat-SLAM} & \textbf{104264} & \textbf{27.69} & \textbf{0.808} & 0.326 \\
    \bottomrule
\end{tabular}}
\vspace{-3mm}
\end{table}
As demonstrated in Table~\ref{tab:ablation_primitives}, the primitives merge strategy in WaterSplat-SLAM reduces memory consumption by 24.3\% while maintaining competitive reconstruction quality. The marginal increase in LPIPS suggests a trade-off between memory efficiency and rendering effects.

\subsection{{Operation Speed}}

\begin{table}[t]
\centering
\caption{Per-frame runtime analysis (second) on WaterSplat-SLAM dataset.}\label{tab:tank_runtime}
\setlength{\tabcolsep}{3pt}
\resizebox{0.48\textwidth}{!}{
\begin{tabular}{lcccccc}
\toprule
Method & Big\_gate & Pipe\_local & Pool\_up & Pool\_up2 & Pool\_loop \\
\midrule
MonoGS \cite{Matsuki_2024_CVPR} & {0.88$\pm$0.24} & {0.84$\pm$0.20} & {1.02$\pm$0.20} & {1.12$\pm$0.37} & {0.67$\pm$0.41} \\
HI-SLAM2 \cite{zhang2024hi} & {0.24$\pm$2.31} & {\textbf{0.17}}$\pm${1.57} & {\textbf{0.24}}$\pm${2.02} & {0.39$\pm$3.33} & {\textbf{0.39}}$\pm${1.74} \\
\textbf{WaterSplat-SLAM} & {\textbf{0.21}}$\pm${0.29} & {0.22$\pm$0.25} & {0.31$\pm$0.37} & {\textbf{0.30}}$\pm${0.43} & {0.47$\pm$1.50} \\
\bottomrule
\end{tabular}}
\vspace{-5mm}
\end{table}

{
The WaterSplat-SLAM system is implemented on a high-performance computing desktop with an NVIDIA RTX 4090 GPU and Intel Core i9-13900K CPU. Our system is compared with HI-SLAM2~\cite{zhang2024hi} and MonoGS~\cite{Matsuki_2024_CVPR} for system operation speed, as they are the only baseline methods that successfully run on all sequences of the WaterSplat-SLAM dataset. As shown in Table \ref{tab:tank_runtime}, our method achieves a stable frame rate ranging from 2 to 5 FPS, which is comparable to HI-SLAM2~\cite{zhang2024hi}, but with significantly smaller deviation. Notably, our method achieves better rendering quality compared to both baselines.}

\section{Conclusion}

We propose WaterSplat-SLAM, a novel monocular underwater SLAM system based on medium-aware 3DGS, achieving robust camera tracking and photorealistic rendering in underwater scenes. The semantic-guided water rendering method achieves high-quality reconstruction of both objects and scattering media. The voxel-based merging of Gaussian primitives upon loop closure reduces redundancy between associated frames, making our medium-aware Gaussian map more efficient. Above all, WaterSplat-SLAM establishes a new foundation for aquatic spatial intelligence systems.

\addtolength{\textheight}{-12cm}

\balance  
\bibliographystyle{IEEEtran} 
\bibliography{reference}     


\end{document}